%% file: Article.tex
\documentclass[times,twoside,onecolumn]{zHenriquesLab-StyleBioRxiv}
\usepackage{blindtext}
\usepackage{booktabs}
\usepackage[normalem]{ulem}
\usepackage{subcaption}
\usepackage{multirow}

\leadauthor{Peiffer} 

\makeatletter
\onecolumn              
\let\twocolumn\relax    
\makeatother

\begin{document}
\title{Portable Biomechanics Laboratory: Clinically Accessible Movement Analysis from a Handheld Smartphone}
\shorttitle{Portable Biomechanics Laboratory}

\author[1,2]{J.D. Peiffer\orcidlink{0000-0003-2382-8065}}
\author[1]{Kunal Shah}
\author[1,3]{Irina Djuraskovic\orcidlink{0009-0008-8940-6790}}
\author[1]{Shawana Anarwala}
\author[1]{Kayan Abdou}
\author[4]{Rujvee Patel}
\author[1,5]{Prakash Jayabalan}
\author[4]{Brenton Pennicooke}
\author[1,5,\Letter]{R. James Cotton\orcidlink{0000-0001-5714-1400}}

\affil[1]{Shirley Ryan AbilityLab, Chicago, IL, USA}
\affil[2]{Department of Biomedical Engineering, Northwestern University, Evanston, IL, USA}
\affil[3]{Interdepartmental Neuroscience, Northwestern University, Chicago, IL}
\affil[4]{Department of Neurological Surgery, Washington University School of Medicine, St. Louis, MO, USA}
\affil[5]{Department of Physical Medicine and Rehabilitation, Northwestern University Feinberg School of Medicine, Chicago, IL, USA}
\maketitle

\input{0abstract.tex}


\begin{corrauthor}
rcotton \{at\} sralab.org

\end{corrauthor}
\section*{Introduction}
\input{1introduction.tex}

\section*{Methods}
\input{2methods.tex}

\section*{Results}
\input{3results.tex}

\section*{Discussion}
\input{4discussion.tex}

\begin{acknowledgements}
\noindent This work was supported by R01HD114776 (RJC), the Restore Center (P2CHD101913), and the Research Accelerator Program of the Shirley Ryan AbilityLab (RJC). JDP is supported by the National Science Foundation Graduate Research Fellowship Program under Grant No. DGE-2234667.
\end{acknowledgements}

\printbibliography

\onecolumn
\newpage

\captionsetup*{format=largeformat}
\setcounter{figure}{0}
\setcounter{table}{0}
\section*{Supplementary Methods}
\input{5supplementary_methods}
\section*{Supplementary Results}
\input{6supplementary_results}

\end{document}

%% file: 0abstract.tex
\begin{abstract}
\noindent
Movement directly reflects neurological and musculoskeletal health, yet objective biomechanical assessment is rarely available in routine care. We introduce Portable Biomechanics Laboratory (PBL), a secure platform for fitting biomechanical models to video collected with a handheld, moving, smartphone. We validate this approach on over 15 hours of data synchronized to ground truth motion capture, finding mean joint‑angle errors < 3° and pelvis‑translation errors of a few centimeters across patients with neurological‑injury, lower‑limb prosthesis users, pediatric in‑patients, and controls. In > 5 hours of prospective deployments to neurosurgery and sports‑medicine clinics, PBL was easy to setup, yielded highly reliable gait metrics (ICC > 0.9), and detected clinically relevant differences. For cervical‑myelopathy patients, its measurement of gait quality correlated with modified Japanese Orthopedic Association (mJOA) scores and were responsive to clinical intervention. Handheld smartphone video can therefore deliver accurate, scalable, and low‑burden biomechanical measurement, enabling greatly increased monitoring of movement impairments. We release the first clinically-validated method for measuring whole-body kinematics from handheld smartphone video at \url{https://IntelligentSensingAndRehabilitation.github.io/MonocularBiomechanics/}.
\end{abstract}

%% file: 1introduction.tex
Many clinical conditions have pronounced movement phenotypes which are rarely measured in clinical practice 
\cite{jacquelin2010gait,balaban_gait_2014,morris_biomechanics_2001}. 
For example, temporal parameters, including stride time and swing time, have been shown to predict fall risk in older adults \cite{hamacher_kinematic_2011}. 
Patients with knee osteoarthritis (KOA) also show altered gait kinematics, which can predict disease progression \cite{heiden_knee_2009, sosdian_longitudinal_2014}.
Gait kinematics can also characterize recovery dynamics after neurological injury and even predict responses to particular interventions \cite{chow_longitudinal_2021,awad_these_2020}.
The \textit{Stroke Recovery and Rehabilitation Roundtable} emphasized the importance of incorporating movement quantification into clinical trials, while noting that logistical challenges in capturing such data remain a significant barrier \cite{kwakkel_standardized_2017}.
Traditionally, gait analysis requires a specialized laboratory equipped with an optical motion capture (OMC) system and force plates, making it expensive, time-consuming, and typically only covered by insurance in limited circumstances \cite{abdullah_multibody_2024}. In some cases, such as orthopedic surgical planning, force plate measurements and electromyography are also required to ensure the precision of marker-based measurements.

Currently, gait is more commonly characterized using the 10-meter walk or 6-minute walk tests, which capture only speed and endurance. Other assessments—such as the Timed Up and Go (TUG) \cite{podsiadlo_timed_1991}, the Berg Balance Scale (BBS) \cite{berg_clinical_1992}, and the Functional Gait Assessment (FGA) \cite{wrisley_reliability_2004}—evaluate movement quality using clinical scoring methods that are either subjective or limited to stopwatch-based timing, but do not directly quantify the movement itself. Scalable, accessible movement analysis in the clinic would substantially improve upon the status quo. 

Recent advances in AI-based methods offer a promising path forward. Multi-view markerless motion capture (MMMC) requires much less time to acquire and less manual post-processing than OMC. In recent years, numerous studies have demonstrated that MMMC produces similar kinematics to OMC in multiple populations \cite{kanko_concurrent_2021, kanko_inter-session_2021, kanko_assessment_2021,riazati_absolute_2022,song_markerless_2023,wren_comparison_2023,horsak_repeatability_2024,dsouza_comparison_2024,wang_evaluation_2025,min_biomechanical_2024}. We have developed methods for MMMC \cite{cotton_biomechanical_2025,cotton_differentiable_2025,firouzabadi_biomechanical_2024,unger_differentiable_2025} using MuJoCo \cite{todorov_mujoco_2012}, a high-performance physics simulator for machine learning that supports differentiable biomechanical models. This enables us to directly end-to-end optimize the kinematics and skeleton scale from videos. Our end-to-end approach outperforms alternatives that reproduce multi-stage marker-based pipelines that first estimate virtual marker locations and then compute inverse kinematics on these trajectories \cite{cotton_differentiable_2025,firouzabadi_biomechanical_2024}. Although MMMC is faster than OMC, installing and calibrating several high-performance cameras remains a significant barrier to clinical implementation.

OpenCap substantially lowers the technical burden of MMMC by using two or three tripod-mounted smartphones, achieving < 5 degrees of joint-angle error relative to OMC \cite{uhlrich_opencap_2023} and detecting movement patterns that classify disease \cite{ruth_video-based_2025,min_biomechanical_2024}. While OpenCap is a significant and widely used contribution, the requirement for two tripods placed at \textpm45 degrees is not always feasible in clinical settings, which often include small exam rooms and tight hallways. In these contexts, a single handheld camera offers a more practical alternative. Monocular video analysis has been used clinically with people with Parkinson’s disease \cite{acevedo_visionmd_2025}, pediatric patients with cerebral palsy \cite{segado_data-driven_2025}, and stroke \cite{lonini_video-based_2022}. However, these approaches infer 2D or 3D joint locations, or non-biomechanically-grounded mesh representations, rather than biomechanical joint angles used in traditional movement analysis. Recent methods \cite{koleini_biopose_2025, xia_reconstructing_2025} have explored data-driven biomechanical reconstruction from a single camera, but have not been trained or validated on clinical populations. To enable data capture in real-world clinical settings, we previously developed a smartphone application for collecting handheld videos in clinics \cite{cimorelli_validation_2024}, which we refer to as the Portable Biomechanics Laboratory (PBL), and validated gait event detection from these recordings \cite{cotton_transforming_2022}. We subsequently demonstrated our end-to-end biomechanical pipeline can reconstruct biomechanically-grounded kinematics from handheld smartphone video, with or without wearable sensors, achieving accurate knee kinematics from video alone \cite{peiffer_fusing_2024}.

\input{figures/design}

The central goal of the present study is to evaluate whether this smartphone-based biomechanics pipeline can produce accurate kinematics across diverse patient populations, movement tasks, and clinical environments that show construct validity against clinical metrics. We perform this validation by first evaluating our system against OMC in control populations, next against MMMC in clinical populations, and finally, our intended use case, collect data in actual clinic hallways to compare against clinical scores (Fig. \ref{fig:design}). Overall, we find PBL has high joint-angle accuracy compared to OMC and MMMC, concordance with clinical scores, and is responsive to clinical intervention. Together, these evaluations demonstrate that quantitative movement analysis can be integrated into routine clinical workflows using a single handheld smartphone for accurate and clinically relevant biomechanical measurement.

%% file: figures/design.tex
\begin{figure}
\centering
\includegraphics[width=1.0\linewidth]{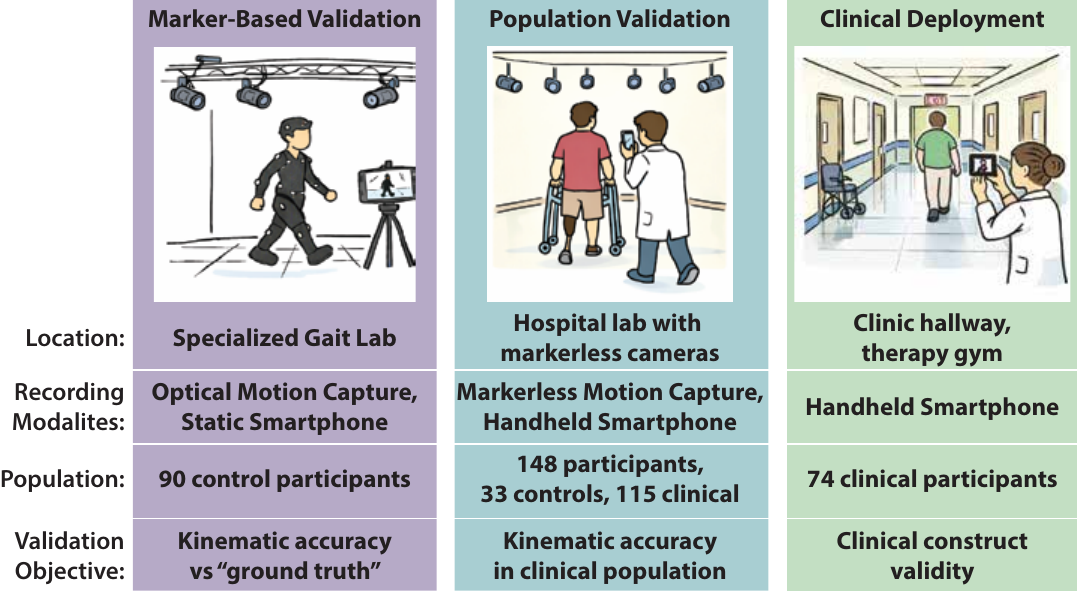}
\caption{\textbf{Study Design Overview.} We first validated the Portable Biomechanics Laboratory (PBL) against an optical motion capture dataset in which control participants performed a range of upper- and lower-limb activities. We then evaluated PBL in clinical populations by recording participants performing lower-limb tasks using both multi-camera markerless motion capture and PBL in hospital laboratories and on clinical floors. Finally, we deployed PBL in clinic hallways and therapy gyms, recording participants after clinical visits to assess the relationship between PBL-derived gait quality scores and standard clinical measures.}
\label{fig:design}
\end{figure}

%% file: 2methods.tex
\subsection*{Portable Biomechanics Laboratory}
We developed a platform called Portable Biomechanics Laboratory (PBL), which records color video, optional depth video, and internal phone sensor information (accelerometer, gyroscopes, and orientation) from a moving, handheld smartphone (Fig. \ref{fig:overview}A). The application can optionally be connected to wearable IMUs with all modalities closely synchronized \cite{cimorelli_validation_2024, peiffer_fusing_2024}. After a recording is finished, PBL uploads the data to a secure cloud platform where data can be pulled into PosePipe \cite{cotton_posepipe_2022}, an open-source package we developed for managing and processing videos.

\input{figures/overview_fig}

\subsubsection*{Video Preprocessing}
Using PosePipe \cite{cotton_posepipe_2022}, we identified the subject of interest in each video using DeepSortYOLOv4 \cite{wojke_simple_2017}. Next, we extracted 2D and 3D virtual marker locations, termed "keypoints", from each video frame using MeTRAbs-ACAE \cite{sarandi_learning_2023} (Fig. \ref{fig:overview}B). We previously identified the 87 keypoints from the MoVi dataset \cite{ghorbani_movi_2021} to be sufficient for full-body kinematics \cite{cotton_optimizing_2023}.

\subsubsection*{Biomechanical Fitting}
We obtained kinematics by optimizing the fit of biomechanical models using our end-to-end differentiable biomechanics framework \cite{peiffer_fusing_2024,cotton_differentiable_2025}. We note that this study estimated joint kinematics only; no inverse dynamics or kinetic estimation was performed. A complete description is in the Supplemental Methods; here we provide a brief overview (Fig. \ref{fig:overview}C,D). We represented a movement trajectory as a neural network that maps time to pelvis location and joint angles, termed an "implicit function". This network was optimized to minimize error between predicted virtual markers and those extracted from video using GPU-accelerated physics simulation in MuJoCo \cite{todorov_mujoco_2012,caggiano_myosuite_2022}.

\subsubsection*{Run Time} Preprocessing a 6 second video takes 10 seconds on a server-grade GPU, 1 minute on a laptop-grade GPU and scales linearly with video length. Biomechanical fitting takes 1 minute on a server-grade GPU, <5 minutes on a laptop-grade GPU, and does not scale with video length.

\subsection*{Datasets}
We applied our monocular fitting approach to three datasets (Fig. \ref{fig:design}). The first is a publicly available dataset that used iPhone videos and an OMC system, but only contained able-bodied individuals. The second of these recorded clinical participants with the PBL system and also an MMMC system. The third dataset reflects our intended use case of videos recorded from participants seen in various clinics, which we used to demonstrate our method can capture clinically meaningful features of gait. Combined, these datasets contain 21 hours of data.

\begin{enumerate}
\item The OMC dataset used was BML-MoVi \cite{ghorbani_movi_2021}, a publicly available dataset of 90 healthy controls performing a variety of sports and upper/lower body movements. It contains 3.8 hours of synchronized video (captured with an iPhone 7) and OMC data using a dense 87-marker set. We fit both modalities of this dataset using our end-to-end differentiable biomechanics approach \cite{cotton_differentiable_2025}, applying the same biomechanical model used for the MMMC and PBL datasets.

\item Our MMMC dataset includes 11.7 hours of recordings from 161 sessions with 148 participants performing walking and lower-limb functional tasks. Of these, 121 participants were recorded simultaneously using both the PBL system \cite{cimorelli_validation_2024} and our MMMC system \cite{cotton_markerless_2023,cotton_differentiable_2025}. This subset included 33 healthy controls, 6 pediatric inpatients, 48 lower limb prosthesis users (LLPUs), and 40 participants with neurological injury. An additional 27 neurological inpatients were recorded using only the MMMC system on a clinical floor. For neurologic patients, we extracted clinical outcome scores—such as the 10 Meter Walk Test (10MWT) and Berg Balance Scale (BBS) \cite{berg_clinical_1992}—from 54 participants across 95 visits. For the BBS, we classify a score of $\leq$ 50 as a fall risk \cite{lusardi_determining_2017} and $\leq$ 40 as high fall risk \cite{shumway-cook_predicting_1997}.

\item Our in-clinic dataset includes 5.5 hours of video from 74 participants recorded with the PBL system during outpatient clinic visits: 55 from a neurosurgery clinic for cervical myelopathy (CM) and 19 from a clinic for knee osteoarthritis (KOA). After their standard physician consultations, participants walked in the clinic hallway while a researcher recorded them using the PBL system. CM participants also completed the Modified Japanese Orthopedic Association (mJOA) questionnaire \cite{kopjar_validity_2011,tetreault_modified_2017}, which scores upper and lower extremity function. Seventeen CM participants were recorded at visits both before and after surgery.

\end{enumerate}
 
\subsection*{Accuracy Evaluation}
To evaluate the accuracy of the smartphone-based fits compared to OMC and MMMC, we computed the Median Absolute Error, defined as the median of joint-angle errors within each trial followed by the median of these per-trial medians across trials. We used Median Absolute Error because both the within-trial and across-trial error distributions were non-Gaussian. When aggregating errors across trials, we report the median and the normalized interquartile range (nIQR), where $nIQR = IQR \cdot 0.7413$.

To compare moving (handheld) vs. static camera video, we also computed monocular fits from single cameras in the MMMC dataset. Because MMMC records participants from all directions, we examined the influence of camera viewpoint on joint-angle error by classifying viewpoint as front, back, ipsilateral, or contralateral.

For static-camera and relevant PBL fits, we report the Root Translation Error (RTE), defined as the average Euclidean error between the pelvis location from the PBL reconstruction and the ground-truth OMC/MMMC reconstruction. For PBL, RTE was computed only for trials involving camera rotation but no camera translation (e.g., TUG trials).

For comparison to OpenCap and other methods, we also compute the Mean Absolute Error following the procedure in \cite{uhlrich_opencap_2023}, detailed in the Supplementary Methods.

\input{figures/overlay_fig}

\subsection*{Clinical Construct Validity of Smartphone-Derived Gait Metrics}
\subsubsection*{Gait Deviation Index}
To reduce kinematic timeseries to an interpretable score representing gait quality, we calculated the Gait Deviation Index (GDI) \cite{schwartz_gait_2008}. Briefly, the GDI uses a dimensionality reduction technique \cite{schwartz_gait_2008,marks_measuring_2018} on cycle-aligned joint kinematics and then measures the distance in this subspace to a distribution of normative gait trials (for which we used 9,008 steps of 206 control participants from both monocular and MMMC systems). We computed GDI, measured between 0 and 100 with 100 being normative gait, on every gait cycle for a participant and then averaged GDI over a session. Finally, we also computed double support time (DST) and cadence using a model for gait event detection that we previously developed and validated \cite{cotton_transforming_2022,cimorelli_validation_2024}.

\subsubsection*{Construct Validity}
Clinical metrics must be repeatable, valid, and responsive. We assessed the repeatability of kinematics from our system by computing intra-class correlations (ICCs) of the GDI between steps taken by the same individual on the same day \cite{bartko_intraclass_1966}. We tested construct validity of our system using t-tests to separate GDI distributions between groups. To further test construct validity, we also compared the GDI to established clinical scores, including the mJOA, 10MWT, and BBS. To evaluate the responsiveness of smartphone-based gait metrics to surgical intervention, we analyzed longitudinal pre- and post-operative changes in mJOA, cadence, DST, and GDI in a cohort of cervical myelopathy participants (n = 17) that were recorded before and after surgery. Paired changes were first assessed for statistical significance using the Wilcoxon signed-rank test. The magnitude and direction of change were then quantified using the corresponding Wilcoxon effect size, which provides a standardized measure on a $[-1,1]$ scale \cite{fritz_effect_2012}. Ninety-five percent confidence intervals for the effect sizes were estimated via bootstrapping. In addition, changes in raw knee kinematics before and after surgery were assessed using paired SPM1D t-tests \cite{pataky_generalized_2010,pataky_one-dimensional_2012}.

%% file: figures/overview_fig.tex
\begin{figure}
\centering
\includegraphics[width=1.0\linewidth]{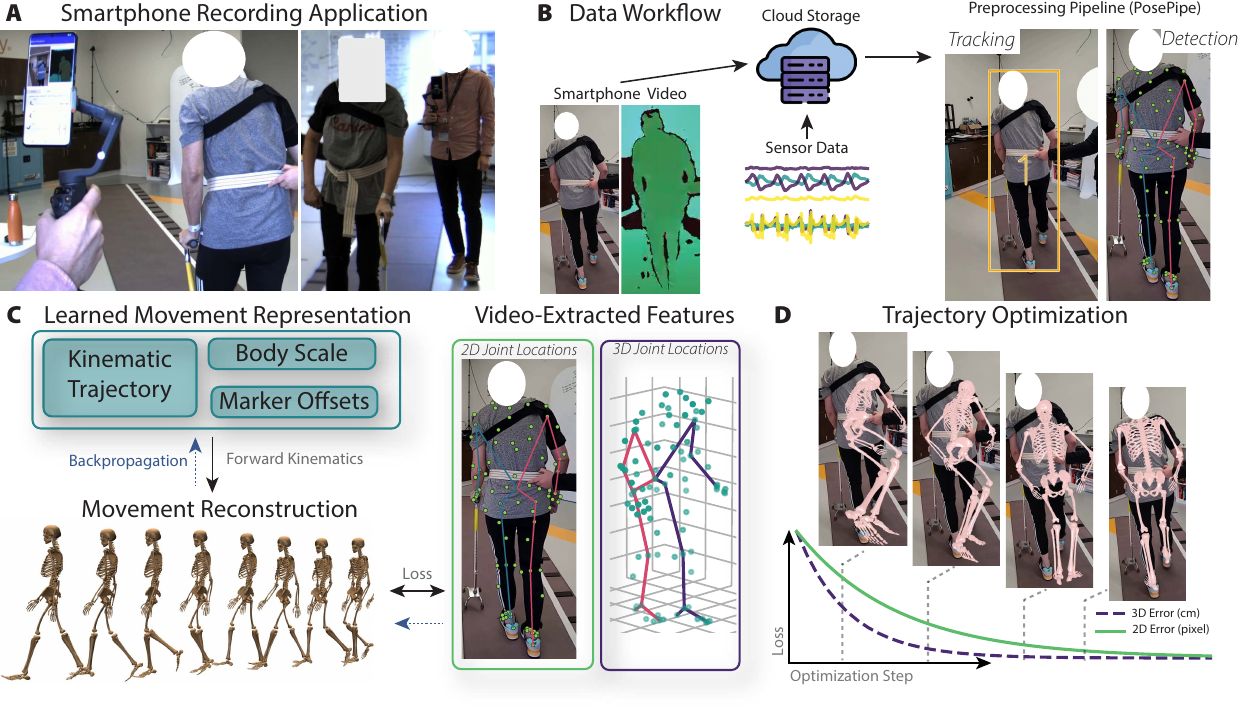}
\caption{\textbf{Biomechanical Reconstruction Overview.} We introduce a method for biomechanically grounded movement analysis in clinical settings using a handheld smartphone. \textbf{A)} Researchers held a smartphone (optionally with gimbal) while following a participant walking. Our system has no specific requirements regarding viewing angle, distance to subject, or therapist assistance. \textbf{B)} Recorded smartphone video and optional wearable sensor data are stored in the cloud, and processed using PosePipe, an open-source package implementing computer vision models for person tracking and keypoint detection. \textbf{C)} To reconstruct movement, we represent movement as a function that outputs joint angles, which—combined with body scaling parameters and evaluated through forward kinematics—generate a posed biomechanical model in 3D space. This untrained model is compared to video-extracted joint locations and optionally smartphone sensor data to compute a loss. This loss guides backpropagation to iteratively refine both the kinematic trajectory and body scale. \textbf{D)} Initially, the representation lacks knowledge of the person’s movements and scale (e.g., height, limb proportions), but after optimization, it typically tracks joint locations within 15 mm in 3D and 5 pixels in 2D.}
\label{fig:overview}
\end{figure}

%% file: figures/overlay_fig.tex
\begin{figure}[!htbp]
\centering
\includegraphics[width=1.0\linewidth]{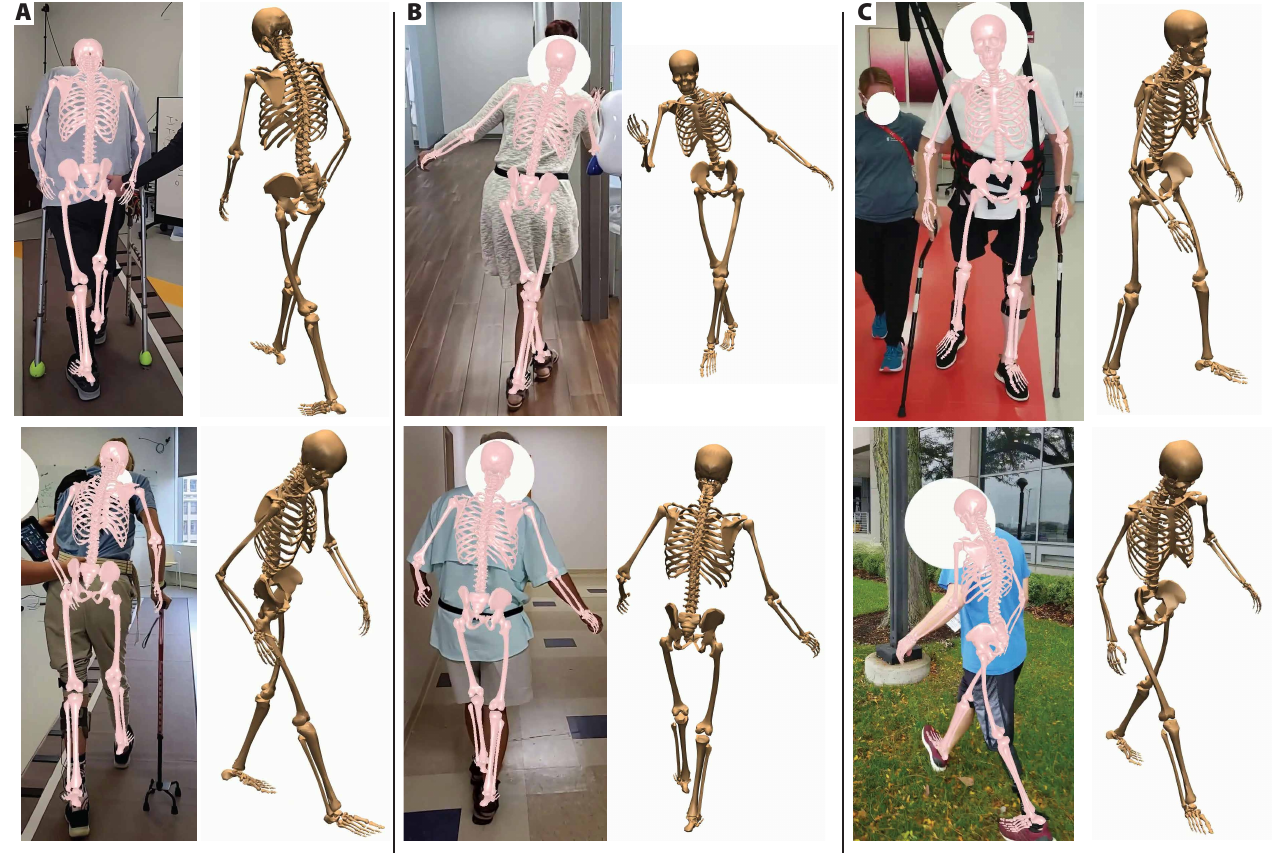}
\caption{
\textbf{Biomechanical Reconstructions from Handheld Smartphone.} Using only a handheld smartphone, our PBL and end-to-end biomechanical fitting approach robustly captures movement across varied clothing types, assistive devices, and clinical environments. \textbf{A)} We first validate this approach on clinical outpatients in our lab, including participants using assistive devices or receiving support from a clinician as needed. \textbf{B)} We next deploy this approach in an outpatient clinic for patients with gait impairments, finding this method minimally disrupting clinical workflow while capturing relevant gait features. \textbf{C)} This technique allows deployable biomechanical capture in dynamic clinical settings as well as outdoors.
}
\label{fig:overlay}
\end{figure} 

%% file: 3results.tex
\subsection*{Ground Truth Comparison}
In general, kinematics from a single camera closely matched MMMC and OMC. Example overlays demonstrate our method is robust to common mobility aids (Fig. \ref{fig:overlay}) and kinematic traces (Fig. \ref{fig:quality}A) reflect clinical impairments like neurologic or prosthetic asymmetries. Quantitative ground truth comparisons are presented in Table \ref{tab:summary} and Supplement.

For the MMMC dataset, the Median Absolute Error between a handheld moving camera and MMMC was 2.79°. This error was slightly higher for static camera fits using a single view from the MMMC system (2.96°), likely due to the greater distances to the person. The error was the greatest in the neurological population, with a Median Absolute Error of 3.32°, lower for LLPUs (2.97°), and lowest for controls at 2.51° (Fig. S\ref{fig:supplementary_histogram}, Table S\ref{tab:monocular_joint_errors_by_population}). 
Our method also performed well on the BML-MoVi dataset, showing only 2.74° of Median Absolute Error from the monocular fits compared to OMC. Error distributions are shown for lower limb joints in Figure \ref{fig:quality}B and broken down across all joints in Tables S\ref{tab:supplementary_joint_errors_by_dataset}, S\ref{tab:monocular_joint_errors_by_population} and Figure S\ref{fig:supplementary_histogram}. Notably wide error distributions are Neck Extension, Lumbar Extension, Hip Flexion (static camera), with third quartiles above 5°.

Frontal-plane joint angles (e.g., hip adduction) were best estimated from frontal views, while sagittal-plane joint angles (e.g., hip and knee flexion) were more accurately estimated from sagittal views (one-sided Mann-Whitney U-tests; p$<$0.005 for all comparisons). This difference was greatest for knee flexion at 1.02° (Fig. \ref{fig:quality}C).

Visual comparisons of root position between single-camera and multi-view fits are shown in Figure S\ref{fig:supplementary_rte}. The clinical dataset using a handheld, rotating smartphone yielded an average RTE of 6.44 cm, while a static camera yielded 5.00 cm. RTE for the BML-MoVi dataset with a static smartphone averaged 2.41 cm (Table \ref{tab:summary}).

\input{tables/summary}
\input{figures/quality_fig}

\subsection*{Clinical Feasibility}
PBL was accessible and easy to use, causing minimal disruption to routine workflows in both KOA and CM clinics. It required no special clothing, lighting, or calibration—we simply followed participants with a handheld phone as they walked.

\input{tables/gdi_cm_summary}

\subsection*{Clinical Inference from Smartphone-Based Kinematics}
Extracting valid, repeatable, and responsive measurements from our smartphone-based system is arguably more important than accuracy alone. We computed the GDI as a measure of gait quality, testing it for repeatability, validity and responsiveness (Figure \ref{fig:gdi_summary}, Table \ref{tab:gdi_cm_summary}). We also test for differences in raw kinematic traces pre/post surgery using the paired SPM1D t-tests \cite{pataky_generalized_2010,pataky_one-dimensional_2012}.

\subsubsection*{Gait Measurement Clinical Validity}
GDI separated groups with expected differences, for instance GDI was significantly lower for participants scoring higher fall risk on the Berg Balance Scale and Transfemoral LLPUs had a significantly lower GDI than Transtibial LLPUs (Fig. \ref{fig:gdi_summary}B,D). The GDI also correlates well with the 10MWT (Pearson's $r = 0.82$), one of the most commonly measured clinical outcomes for gait (Fig. \ref{fig:gdi_summary}C). Finally, the GDI showed a significant correlation with mJOA ($r=0.47$) and the lower extremity subscore of the mJOA ($r=0.48$) (Fig. \ref{fig:gdi_summary}E, Table \ref{tab:gdi_cm_summary}).

\subsubsection*{Gait Measurement Repeatability}
We analyzed the $>$13,500 steps across 222 individuals recorded with the PBL system, and found high repeatability of 0.84 for the GDI of a single step (ICC2) and higher repeatability when averaging multiple steps together: 0.96 (ICC2k) (Table \ref{tab:gdi_cm_summary}). ICCs for spatiotemporal metrics (cadence, DST) were lower-likely due to timing variability—but remained highly repeatable when averaged (ICC2k).

\subsubsection*{Responsiveness of Gait Measurement to Clinical Intervention}
Smartphone-derived gait metrics (GDI and DST) demonstrated significant postoperative changes in individuals with cervical myelopathy (Wilcoxon signed-rank test, p<0.05), whereas patient-reported clinical scales (mJOA and mJOA-LE) did not show significant change (Table \ref{tab:gdi_cm_summary}). Wilcoxon effect sizes were 0.24 for mJOA, 0.22 for mJOA-LE, -0.52 for DST, and 0.44 for GDI. In addition, paired SPM1D t-tests revealed a significant postoperative increase in knee flexion during initial swing (64–68 \% of the gait cycle).

%% file: tables/summary.tex
\begin{table}
\vspace{1.5em}
\centering
\caption{\textbf{Summary of errors for different datasets.} Median and Mean Absolute Errors are for joint angles aggregated across all joints, participants, and trials. RTE is the median root translation error (in cm). Errors are presented as: (Median (nIQR)).}
\label{tab:summary}
\begin{tabular}{lccc}
\toprule
 & \multicolumn{2}{c}{MMMC} & OMC \\
 & Handheld & Static & Static \\
\midrule
Median Absolute Error (deg) & 2.79 (0.86) & 2.96 (0.78) & 2.74 (0.50) \\
Mean Absolute Error (deg) & 4.10 (3.71) & 4.49 (1.99) & 3.89 (1.66) \\
RTE (cm) & 6.44 (4.77) & 5.00 (4.68) & 2.41 (0.63) \\
\bottomrule
\end{tabular}
\vspace{-1em}

\end{table}

%% file: figures/quality_fig.tex
\begin{figure}
\centering
\includegraphics[width=1.0\linewidth]{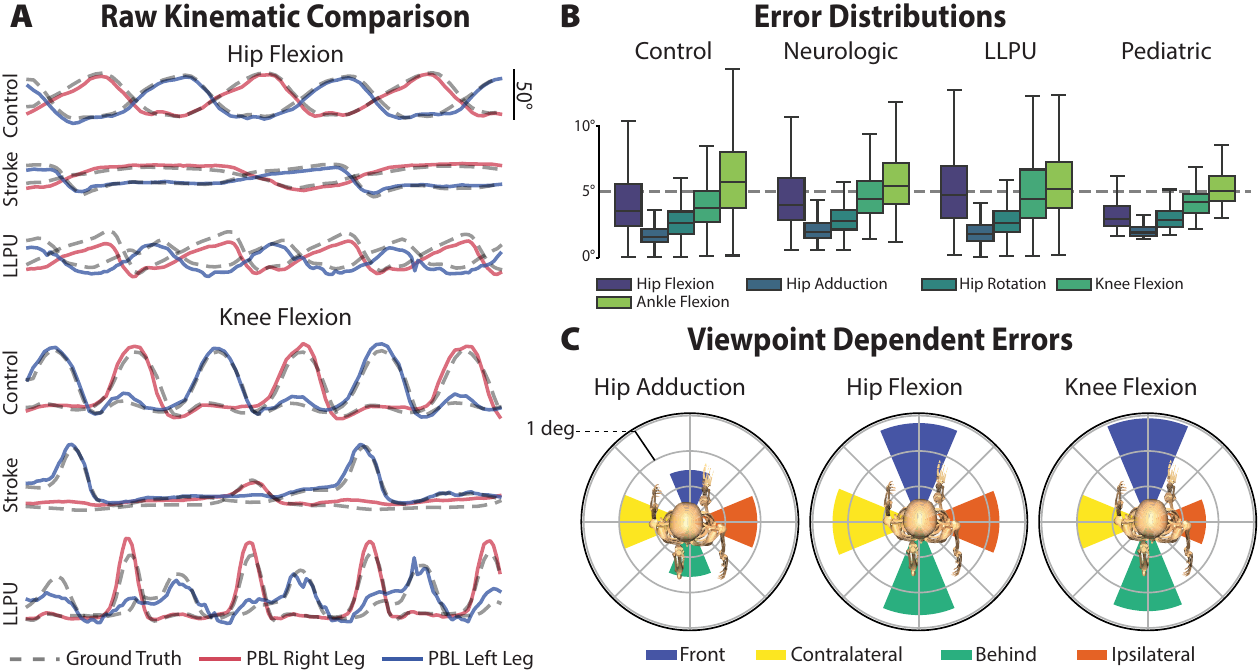}
\caption{\textbf{Quality Measures of Single Camera Fitting.} \textbf{A)} Kinematic traces from smartphone video (red/blue) compared to ground truth (gray dashed) during walking. \textbf{B)} Joint angle errors across populations for select lower limb angles. Control: n=29 participants, v=950 videos, Neurologic: n=36, v=290, LLPU: n=40, v=458, Pediatric: n=4, v=25. \textbf{C)} Select joint angle errors with respect to camera view angle show that sagittal plane angles have the lowest error with sagittal camera views, and frontal angles have the lowest error with frontal views.}
\label{fig:quality}
\end{figure}

%% file: tables/gdi_cm_summary.tex
\begin{table}[!hbtp]
\vspace{1em}
\centering
\caption{\textbf{Reliability, Validity, and Responsiveness of Smartphone-Based Metrics.} ICCs are presented as ICC (95\% confidence intervals). mJOA and mJOA-LE show Spearman correlations between gait metric and ordinal ratings. SRM a unitless measure of change in metric before and after surgery, with 95\% confidence intervals. Wilcoxon Effect Size presented with 95\% confidence intervals. * indicates $p<0.05$.}
\label{tab:gdi_cm_summary}
\begin{tabular}{ll|cc|cc|c}
\toprule
 & & \multicolumn{2}{c}{\textbf{Repeatability}} & \multicolumn{2}{c}{\textbf{Validity}} & \textbf{Responsiveness} \\
 & & ICC2 & ICC2k  & r mJOA & r mJOA-LE  & Effect Size \\
\midrule
 \multirow{2}{*}{Patient-Reported}
 &mJOA & - & -  & - & -& 0.24 (-0.35 0.76) \\
 &mJOA-LE & - & -  & - & - &  0.22 (-0.41 0.91)\\
\midrule
 \multirow{3}{*}{Smartphone-Based}
 &Cadence & 0.58 (0.52 0.64) & 0.88 (0.85 0.90)  & 0.25* & 0.39* & 0.39 (-0.09 0.77)\\
 &DST & 0.41 (0.35 0.48) & 0.78 (0.72 0.82) & -0.27* & -0.37* & -0.52 (-0.84 -0.11)*\\
 &GDI & 0.84 (0.81 0.86) & 0.96 (0.96 0.97)  & 0.47* & 0.48* & 0.44 (-0.02 0.76)*\\
\bottomrule
\end{tabular}
\end{table}

%% file: 4discussion.tex
\input{figures/gdi_fig}

We present a clinically validated method for extracting biomechanically accurate kinematics from smartphone video in clinical settings. Across diverse populations, our single-camera system showed strong agreement with ground truth (typically $<3^\circ$ Median Absolute Error). Gait quality metrics derived from these videos detected clinically relevant group differences and were responsive to clinical intervention. These findings suggest that a clinician, therapist, or medical assistant could feasibly perform gait analysis in under a minute to obtain an objective measurement of a patient's gait impairment to track their recovery.

Joint angle errors were generally low across all populations and settings (Table \ref{tab:summary}). We attribute the slightly higher Median Absolute Error in clinical populations (0.46 - 0.81 degrees, Fig. \ref{fig:quality}B, Fig. S\ref{fig:supplementary_histogram}B, Table S\ref{tab:monocular_joint_errors_by_population}) to assistive devices such as canes, walkers, ankle foot orthoses, or gait belts (Fig. \ref{fig:overlay}) and the fact that keypoint detection does not perform as well on prosthetic and orthotic devices \cite{cimorelli_validation_2024}. Studies using OpenCap also report increases in joint angle errors in clinical populations \cite{wang_evaluation_2025}. Intuitively, sagittal-plane angles were most accurate when recorded from sagittal viewpoints, and frontal-plane angles were most accurate from frontal viewpoints (Fig. \ref{fig:quality}C), giving clinicians a simple and actionable strategy for mitigating joint-angle error during data collection.

Our only dataset comparing PBL to Optical Motion Capture (OMC) is BML-MoVi, which only contains able-bodied controls. To further validate the biomechanical generalization of PBL, we also included a dataset with diverse etiologies of gait impairments assessed with our previously validated MMMC system \cite{cotton_markerless_2023, cotton_differentiable_2025}. Although MMMC differs from OMC, multiple independent studies have demonstrated strong agreement between the two approaches \cite{kanko_concurrent_2021, kanko_assessment_2021, riazati_absolute_2022, song_markerless_2023, wren_comparison_2023, horsak_repeatability_2024, dsouza_comparison_2024, unger_differentiable_2025}. Nonetheless, neither method constitutes a true biomechanical gold standard relative to biplanar fluoroscopy, due to known limitations including marker placement variability \cite{uchida_conclusion_2022} and soft-tissue artifact \cite{akbarshahi_non-invasive_2010}. The consistently strong correspondence observed between PBL and both OMC and MMMC across a diverse participant population supports the biomechanical accuracy and generalizability of the proposed approach (Table \ref{tab:summary}). While monocular video-based analysis is expected to exhibit some loss of accuracy relative to multiview systems, it substantially improves accessibility in clinical environments. Thus, the latter part of this study evaluates whether PBL collected in clinical settings is sufficiently accurate to demonstrate construct validity against established clinical metrics.

To demonstrate this, we showed that movement quality scores from PBL: (1) change with expected population changes, (2) correlate with clinical scores, and (3) are responsive to clinical intervention. First, GDI from PBL was decreased for inpatients with fall risk assessed by the BBS and further decreased for inpatients classified as high fall risk (Fig. \ref{fig:gdi_summary}B). Similarly, GDI was decreased for Transtibial LLPUs and further decreased for Transfemoral LLPUs, an intuitive result. Second, GDI correlated highly with the 10MWT (Fig. \ref{fig:gdi_summary}C), a widely used clinical outcome measure. The GDI also had a moderate correlation with the mJOA (Fig \ref{fig:gdi_summary}E, Table \ref{tab:gdi_cm_summary}), which is expected considering the GDI is a continuous quantitative measure and the mJOA is a patient-reported ordinal scale. Regarding responsiveness, SPM1D analysis of kinematic traces showed significant changes in knee flexion following surgical intervention, establishing that our system can be used to analyze specific changes in movement and quantitatively track recovery. Quantitative metrics (DST and GDI) extracted from our smartphone-based system showed significant postoperative changes, whereas the mJOA did not (Table \ref{tab:gdi_cm_summary}). This may suggest that smartphone-derived kinematic measures capture postoperative kinematics changes not reflected in patient-reported scores, but it is important to highlight these are measuring fundamentally different constructs. Thus, quantitative gait metrics (e.g., GDI, DST) complement patient-reported outcomes such as the mJOA, and we defer to future studies to investigate whether they may precede and causally influence patient outcomes \cite{cotton_causal_2024}.

We focused on the GDI because it is a widely used summary of gait kinematics. However, we do not claim it is the optimal metric for every application. Instead, we anticipate that scalable gait analysis will enable new research to identify the most clinically meaningful measures. For example, we recently described a method for inferring torques and ground reaction forces from kinematics and saw that the ground reaction forces are sensitive to clinical history \cite{cotton_kintwin_2025}.

The minimally clinically important difference for kinematic features of gait has not yet been established for many clinical conditions. While our accuracy is generally quite high, greater accuracy may be required for certain indications. For example, the 5 degrees of error we typically see at the ankle might be insufficient to detect subtle changes in plantarflexion spasticity or toe clearance from a single camera.

We observe heavy-tailed error distributions, with some trials exceeding 10° of error (Fig. S\ref{fig:supplementary_histogram}C). Although occasional large errors and failure cases warrant further investigation, these errors occur infrequently and do not introduce systematic bias. Aggregated kinematic trajectories and summary metrics remain strongly correlated with clinically relevant measures (Table \ref{tab:gdi_cm_summary}, Fig. \ref{fig:gdi_summary}), supporting the method’s reliability for population-level analysis and leaving subject-level clinical decision-making to future work. We tested whether fitting residuals—2D reprojection error or 3D keypoint error—could serve as reliability metrics. Their correlations with joint-angle error were weak (0.35 and 0.24, respectively; Fig. S\ref{fig:supplementary_residuals}), underscoring the need for better confidence estimates in markerless motion capture. We have previously developed calibrated confidence intervals for monocular pose estimation without biomechanics \cite{pierzchlewicz_platypose_2024} and for multi-view biomechanical estimation \cite{cotton_biomechanical_2025}; extending these ideas to monocular biomechanics is an important direction for future work.

Other approaches have proposed deployable systems for human motion capture. Most well-known is OpenCap \cite{uhlrich_opencap_2023}, which uses two or three smartphones on static tripods to triangulate joint locations from video and fit biomechanics using the widely used OpenSim framework \cite{delp_opensim_2007,werling_addbiomechanics_2023}. OpenCap reports a Mean Absolute Error of 4.1° during walking in 10 controls using two static cameras. In our control cohort (n=29), PBL achieves a comparable Mean Absolute Error of 4.32 (2.06)° using a single handheld moving camera. While this error was computed on similar populations and activities, we note this is not a direct comparison and leave that to future work. While more cameras will generally increase accuracy \cite{cotton_biomechanical_2025}, our results suggest that the difference between OpenCap with 2 camera on tripods and a single handheld smartphone is now small. Apart from accuracy, OpenCap and our PBL have different advantages: OpenCap provides a very user friendly platform, uses the widely accepted OpenSim package \cite{delp_opensim_2007}, and calculates joint kinetics. Our PBL approach presented here is markedly easier to deploy in clinical settings such as narrow hallways, small exam rooms, and busy gym spaces. The same simplicity also enables potential use in patients’ homes with recordings performed by caregivers. Extending our method to include joint kinetics is feasible with future work such as \cite{cotton_kintwin_2025}. While our pipeline is not real-time, per-trial runtimes are compatible with same-day or visit-level clinical analysis.

Two recent approaches estimate biomechanics from monocular video: BioPose \cite{koleini_biopose_2025} and Human Skeleton and Mesh Recovery (HSMR) \cite{xia_reconstructing_2025}. Both regress pose directly from images, making them faster than our pipeline. However, neither was trained or evaluated on clinical populations, so their performance on clinical gait—often far outside the distribution of healthy movement (Fig. \ref{fig:gdi_summary}A)—is uncertain. Our method instead uses trajectory optimization with in-loop biomechanical model scaling applied consistently across trials, aligning more closely with established biomechanical reconstruction practices. It can also integrate wearable sensors \cite{peiffer_fusing_2024} to address occlusion and high-speed motion. For healthy controls, our kinematic accuracy is similar to BioPose (<3.9° Mean Absolute Error on MoVi dataset), though the comparison is incomplete because BioPose does not specify which joints contribute to its reported errors. HSMR does not report joint-angle errors. Ultimately, broader development of AI-driven biomechanics tools will expand clinical movement analysis, and large resulting datasets will support data-driven, precision rehabilitation strategies \cite{cotton_causal_2024}.

In conclusion, we present a method for extracting accurate and biomechanically interpretable joint kinematics from smartphone videos. We validate this method in inpatient, outpatient therapy, and clinical settings, finding it to be highly usable and robust—even in the presence of nearby therapists and assistive devices. Importantly, gait scores extracted from this approach correlate well with commonly used clinical scales such as the 10MWT, BBS, and mJOA, and respond to clinical intervention. This approach enables increased fidelity in clinical and at-home monitoring of movement impairments, paving the way for a big data revolution in movement science.

%% file: figures/gdi_fig.tex
\begin{figure}
\centering
\includegraphics[width=0.95\linewidth]{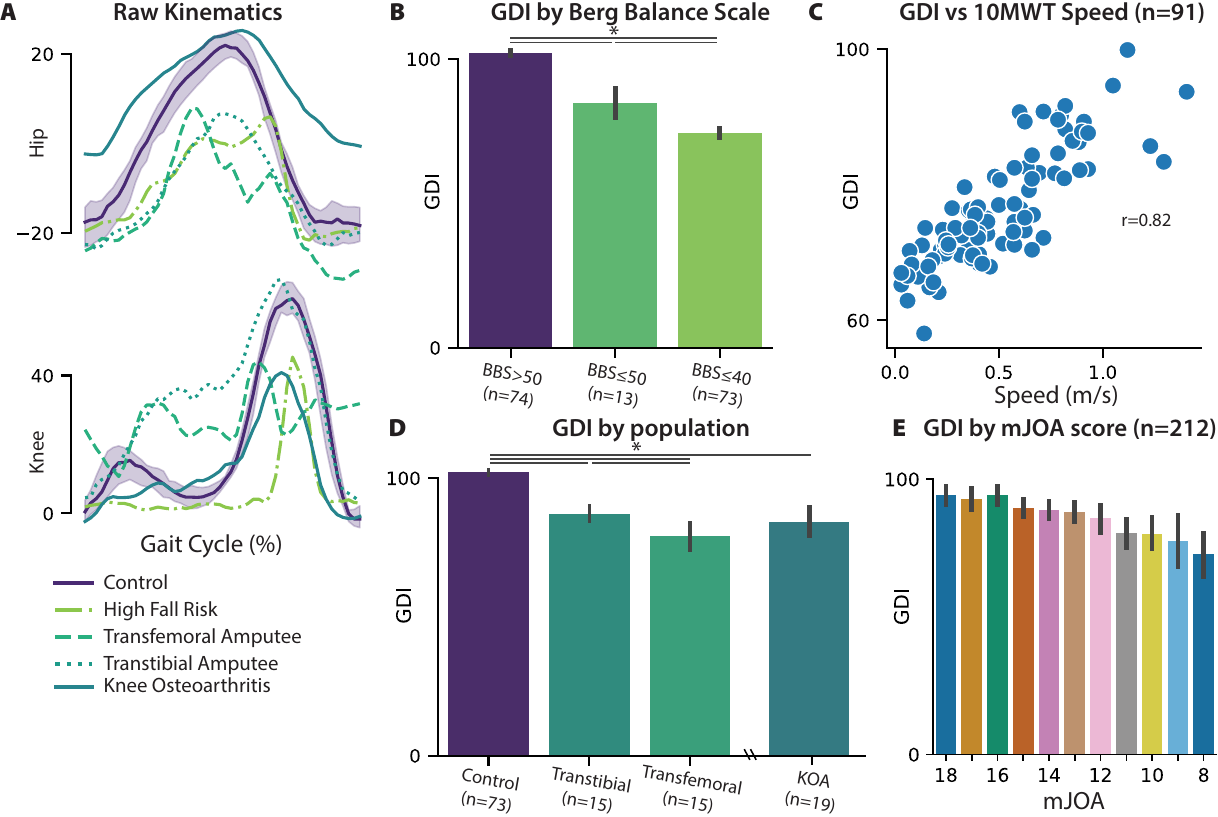}
\caption{\textbf{Clinical Validity of Smartphone-Based Gait Deviation Index.} \textbf{A)} Hip and Knee flexion angles of clinical and control groups \textbf{B)} GDI separates groups at risk of falls determined by the Berg Balance Scale. \textbf{C)} GDI correlates with 10 Meter Walk Test performance $r = 0.82$. \textbf{D)} GDI of LLPUs and KOA participants is significantly lower than that of control populations. Further, GDI of Transfemoral amputees is significantly lower than GDI of Transtibial amputees. \textbf{E)} GDI collected in clinical settings correlates ($r = 0.47$) with the mJOA, a clinically used ordinal questionnaire.}
\label{fig:gdi_summary}
\end{figure}

%% file: 5supplementary_methods.tex
\subsection*{Datasets}
\subsubsection*{OMC Dataset: BML-MoVi}
The publicly available BML-MoVi \cite{ghorbani_movi_2021} dataset contains synchronous optical motion capture (OMC) data and smartphone RGB video data of 90 able-bodied participants performing a variety of everyday and sports movements. The OMC data contains a dense set of 87 markers, suitable for reconstructing biomechanics recorded at 120 Hz.

Smartphone video was collected at 30 Hz and 1080 x 1920 resolution using an iPhone 7. The smartphone video totals 415,060 frames for 3.8 hours of video.

For the OMC data, we fit our model using only the 3D keypoint loss as the markers had already been triangulated and cleaned, following the approach from \cite{cotton_differentiable_2025}. For the smartphone data, we apply the 3D and 2D keypoint loss following the method we use on the PBL data and describe below. 
\subsubsection*{MMMC Dataset}
This study was approved by the Northwestern University Institutional Review Board. Participants provided informed consent for their videos to be stored, processed, and de-identified videos to be shared in publication. Data were collected in our laboratory using handheld smartphone video \cite{cimorelli_validation_2024}, an instrumented GaitRite walkway, and an MMMC system \cite{cotton_markerless_2023}. The dataset includes recordings from 33 control participants, 48 lower limb prosthesis users (LLPUs), and 73 individuals with a history of neurological injury, including 6 pediatric cases. All recording sessions consisted of overground walking along the GaitRite walkway (8 m) while a researcher followed the participant with smartphone.

Control participants typically completed three recording sessions. The first session included nine walking trials, while subsequent sessions incorporated additional assessments such as the Four Square Step Test (FSST), Postural Sway Test (PST), Timed Up and Go (TUG), and Tandem Gait, averaging 15 recordings per session. LLPUs were generally recruited from an outpatient clinic for a single session including the additional tasks if the participant felt comfortable. The neurologic participants were largely outpatients (40) collected with both PBL and MMMC. We also analyzed single-view videos from MMMC data collected from 27 participants admitted to inpatient rehabilitation using static cameras installed in therapy gyms at Shirley Ryan AbilityLab. These participants walked at a self selected speed using whatever assistive device they generally used (e.g. walker, cane, ankle-foot orthosis, gait belt). 

\textit{Portable Biomechanics Laboratory (PBL)}
All smartphone RGB video was captured on a Samsung Galaxy S8 or Samsung Galaxy S20 Ultra at 30 Hz and 1080 $\times$ 1920 resolution \cite{cimorelli_validation_2024} using a custom application. This application simultaneously logs smartphone RGB video, and if available, depth video, internal phone sensor logs (gyroscope, magnetometer, accelerometer, and fused orientation estimate), and wearable sensor data \cite{cotton_wearable_2019}. During walking trials, a researcher held the phone in portrait orientation mounted on a 3-axis gimbal and followed the participant from behind at a mostly sagittal, but partially oblique view. The gimbal was used solely to improve visualizations; the method remains accurate without it, though the resulting video may be slightly shakier. FSST was collected at a frontal view in front of the participant, and TUG was collected from a sagittal view with the camera rotating to follow the participant.

The video and sensor data from the smartphone application was uploaded to a secure cloud server for storage and later downloaded into PosePipe \cite{cotton_posepipe_2022}, an open source application implementing a variety of computer vision tasks such as person detection, annotation, and 2D and 3D keypoint detection. PosePipe tracks all data and analyses in a MySQL database using DataJoint \cite{yatsenko_datajoint_2015}. This makes it feasible to manage the 100s of thousands of videos we have collected.

\textit{Multi-View Markerless Motion Capture (MMMC)}
Recordings in the laboratory dataset employed the Multi-View Markerless Motion Capture (MMMC) system which consists of 8-12 FLIR BlackFly S GigE cameras which acquire synchronized RGB video at 29 fps. Cameras were arranged such that at least three cameras covered participants at all points during a recording. A detailed description of this system can be found in \cite{cotton_markerless_2023}. For this system, we reconstructed biomechanical fits using methods described in \cite{cotton_differentiable_2025}. In total, the dataset includes 1,267,586 frames across 2,206 trials from 143 participants—comprising 1,056,861 frames from 2,028 trials of 116 participants recorded with both the PBL and MMMC systems, and an additional 210,725 frames from 178 trials of 27 participants recorded with a single MMMC camera—amounting to 11.7 hours of video.

\subsubsection*{In-Clinic Dataset}
This study was approved by the Northwestern University and Washington University in St. Louis Institutional Review Boards. Participants provided informed consent for their videos to be stored, processed, and de-identified videos to be shared in publication. To validate our methods in a real-world clinical environment, we deployed our PBL system \cite{cimorelli_validation_2024} in two outpatient clinics. The first was a neurosurgery clinic treating patients with cervical myelopathy (CM)—a degenerative condition of the cervical spine that often impairs gait and balance. During routine pre- and post-operative visits for spinal decompression, participants were asked to walk three times at a self-selected pace and three times at a fast pace down the clinic hallway while being recorded from behind by a clinician. At each visit, patients completed the modified Japanese Orthopedic Association (mJOA) questionnaire \cite{yonenobu_interobserver_2001,kato_comparison_2015} to assess their symptoms. The CM cohort includes 911 videos from 55 patients collected at multiple timepoints: pre-surgery, 6 weeks post-surgery, 3 months post-surgery, and 1 year post-surgery. 17 of these participants have sessions before and after surgery, which were used in our sensitivity analysis. 

The second cohort consisted of 110 videos of 19 participants with knee osteoarthritis (KOA) coming to clinic for corticosteroid injection. As before, these videos were filmed in the hallway before or after the visit.

In total, this dataset contains 589,557 frames or 5.5 hours of video in clinical settings.

\subsection*{Keypoint Detection}
We employed MeTRAbs-ACAE \cite{sarandi_learning_2023} to detect 2D and 3D virtual marker locations from RGB videos. As recommended by the author, we approximated confidence by measuring the standard deviation of each 3D joint location estimated from 10 different augmented versions of each video frame. This was converted to a confidence estimate using a sigmoid function with a half maximum at 30 mm and a width of 10 mm. This results in a confidence score $c(t,j) \in [0,1]$ for each timepoint $t$ and joint $j$. For training, we only used frames where the person was fully in the video view, allowing for short periods ($<$1.0) seconds of partial coverage, reducing total dataset size to 12.4 hours. For a 6 second video with one person in frame, MeTRAbs-ACAE took 10 seconds to run on a NVIDIA A100 GPU and 1 minute to run on a NVIDIA GeForce RTX 3050 Laptop GPU. This processing time will scale linearly with video time.

\subsection*{Differentiable Biomechanical Model}
Smartphone and MMMC reconstructions utilized a biomechanically-grounded model implemented in Mujoco from LocoMujoco \cite{al-hafez_locomujoco_2023}, originally based on an OpenSim model \cite{hamner_muscle_2010}. Mujoco's GPU acceleration engine, MJX, supports parallelizing forward kinematic passes of this model. We have previously optimized site locations for the 87 MoVi keypoints on a similar model \cite{cotton_optimizing_2023}. We also removed collisions other than the feet, added a neck joint with 3 degrees of freedom, and extended the hip flexion and extension ranges. Model scaling is controlled by 8 scaling parameters: overall size, the pelvis, left thigh, left leg and foot, right thigh, right leg and foot, the left arm, and the left leg. We follow notation used by the skinned multi-person linear model (SMPL) \cite{loper_smpl_2015} and represent our GPU-accelerated, biomechanically-ground, forward kinematic equation as:
\begin{equation}
\mathbf{x} = \mathcal{M}(\theta,\beta)
\label{eq:fk}
\end{equation}

\noindent where $\theta \in \mathbb{R}^{40}$ are joint angles that pose the model, $\beta \in \mathbb{R}^{8+87 \times 3}$ are the 8 scaling parameters and an offset for each of the 87 keypoints. $\mathbf{x} \in \mathbb{R}^{87 \times 3}$ are the marker locations following the scaling, site offset and forward kinematics.

\subsection*{Implicit Representation}
We represent the kinematic trajectory for each trial as a learned implicit function, $f_\theta$, implemented as a multi-layer perceptron that takes time as an input using sinusoidal positional encoding \cite{vaswani_attention_2017} and outputs joint angles $\theta(t)$. In addition to the 40 joint angles output from the final layer, we also output 3 parameters representing the orientation of the smartphone in global space $\mathbf{r}$.

\begin{equation}
f_\phi: t \rightarrow (\mathbf{\theta}, \mathbf{r})
\label{eq:implicit}
\end{equation}

\noindent The outputs of $f_\phi$ corresponding to rotations (i.e., not the pelvis location) are passed through a tanh nonlinearity to limit it to $(-1,1)$ followed by scaling this range to match the biomechanical model joint limits. Joint angles, $\theta(t)$, are then passed through the forward kinematic equation above (Eq. \ref{eq:fk}) to obtain the 3D joint locations, $\mathbf x (t)$, at that time.

\subsection*{Reference frames} We define the global reference frame $\{n\}$ for our model output. Since the PBL system tracks changes in phone orientation from the smartphone IMUs, we are able to relate the orientation of the camera frame $\{c\}$ to the orientation of the world frame at any point in time using $R_{nc} \in SO(3)$. We follow the convention presented in \cite{lynch_modern_2017} for rotations where $R_{ab}$ represents frame $\{b\}$ relative to frame $\{a\}$.

\subsection*{Model Fitting}
Each trial is represented as a unique implicit function. We jointly optimize the parameters of the implicit representations for N trials $\{ \phi_0, \phi_1, ..., \phi_{N-1} \}$, the body scaling, and site offset parameters $\beta$, to follow 2D and 3D keypoints extracted from video as well as orientation data recorded from the smartphone. This approach of jointly learning kinematics and body scaling and marker offset for the entire session has been called bilevel optimization \cite{werling_addbiomechanics_2023}. For example, a session consisting of 10 trials would jointly optimize 10 implicit functions $f_{\theta_i}$ and a single set of body scaling parameters $\beta$. A single trial required 1 minute to optimize on a NVIDIA A100 GPU and 4.5 minutes to optimize on a consumer-grade (NVIDIA GeForce RTX 3050 Laptop) GPU. Trial length has little effect on runtime, and additional trials add only an incremental processing cost.

For each training step, we evaluated every implicit function with a batch of 300 time samples for each trial to extract the predicted joint angles at those time points $\hat \theta(t)$, which is performed in parallel across all trials. Using these poses and scaling parameters, we performed GPU accelerated forward kinematic passes of our model (Eq. \ref{eq:fk}) to predict 3D joint locations for that training batch, $\hat{\mathbf x} = \mathcal M(\hat \theta(t), \hat{\beta})$. Next, we describe the losses used during this optimization. To reduce notational clutter, we drop the explicit time and trial dependence in the loss definitions.

\subsubsection*{3D Keypoint Loss}
We obtain a pure video estimate of the 3D keypoint locations in the camera reference frame $x_c$ using MeTRAbs-ACAE \cite{sarandi_learning_2023}. We can rotate these keypoints into the global frame using $x_n=\hat R_{nc}x_c$ and define a loss function on the Euclidean distance in centimeters between keypoints from the model $\hat x_n$ with the video keypoints as:

\begin{equation}
\mathcal L_{3D}(\phi,\vec \beta)=\frac{1}{J}\sum_{j\in J} c(j) g(\Vert \hat x_n - \hat R_{nc}x_{c} \Vert_2)
\label{eq:3D loss}
\end{equation}

\noindent where $c(j) \in [0,1]$ is the confidence score for joint keypoint $j$, which also varies with time, and $g(\cdot)$ is a Huber loss which is quadratic within 10 cm and linear after, necessary for stabilizing early training. This loss is computed between the 3D keypoints set with their mean translation removed.

\subsubsection*{2D Keypoint Loss}
MeTRAbs-ACAE also produces 2D keypoints in the image frame $\vec u \in \mathbb{R}^{87 \times 2}$. The 3D keypoints produced by our model $\hat x_n$ can be rotated into the camera frame using $\hat R_{nc}$ and projected through a camera model $\Pi$ with the calibrated camera intrinsics, to compute the error in pixels with detected 2D keypoints:

\begin{equation}
\mathcal L_{2D}(\phi,\vec \beta)=\frac{1}{J}\sum_{j\in J} c(j)g( \Vert \Pi(\hat R_{nc}^{-1}\hat x_n) - \vec u \Vert_2)
\label{eq:2D loss}
\end{equation}

In this loss, $g(\cdot)$ is a Huber loss which is quadratic within 5 pixels and linear after, which reduces the sensitivity of the fits to outliers.

\subsubsection*{Phone orientation loss}
For videos collected with our PBL platform that include phone orientation data, we evaluate the implicit functions at the phone IMU sample points and extract the predicted phone orientation trajectory $\hat R_{nc}(t)$. We found representing $R_{nc}$ as a rotation vector led to more stable training; however, the phone orientation output from internal sensors was represented as a quaternion, so we convert $R_{nc}$ to a quaternion for calculating the angular difference between the predicted and measured orientation, measured in degrees with the following loss:

\begin{equation}
\mathcal{L}_{phone} =\frac{180}{\pi}\cdot 2 \cdot \arctan\left(\frac{\sqrt{q_1^2 + q_2^2 + q_3^2}}{|q_0|} \right)
\end{equation}

\subsubsection*{Total loss and optimization}

These terms were combined with hyperparameters ($\lambda_1 = 1, \lambda_2=1e-1,\lambda_3=1$), to control their relative weights and provide an overall loss:

\begin{equation}
\mathcal L = \lambda_1 \mathcal L_{3D}+ \lambda_2\mathcal L_{2D} + \lambda_3 \mathcal L_{phone}
\end{equation}

We defined implicit functions and computed the loss using Equinox \cite{kidger_equinox_2021} and JAX \cite{deepmind2020jax}, optimizing with the Adam optimizer for 25,000 iterations of gradient descent. Weight decay of $1\mathrm{e}{-5}$ was applied to the implicit function parameters. With Optax, we set the learning rate to start at $1\mathrm{e}{-3}$ and decay exponentially to $1\mathrm{e}{-6}$ \cite{kingma_adam_2017,deepmind2020jax}.

For MMMC \cite{cotton_differentiable_2025}, we found random initialization of the network sufficient for optimizing implicit representations. However, for monocular videos, this often failed to converge if the subject wasn't initially visible to the camera. To address this, we adjusted each trial's implicit function by biasing the final layer to place the pelvis 1.5 meters from the camera and initializing $R_{nc}$ outputs to the median of observed $R_{nc}$ values.

\subsection*{Validation Metrics}
We evaluated normality of errors at two levels using the Shapiro–Wilk test \cite{shapiro_analysis_1965}: (1) the distribution of joint angle errors within each trial and (2) the distribution of trial-level error summaries across trials. Both levels were non-Gaussian. Consequently, we report the Median Absolute Error, computed by first taking the median joint-angle error within each trial and then taking the median of these per-trial medians across trials. The Median Absolute Error includes the following degrees of freedom: hips (2×3), knees (2×1), ankles (2×1), lumbar spine (3), neck (3), shoulders (2×2), and elbows (2×1).

When aggregating results across trials, we also report the normalized interquartile range, $nIQR = IQR \cdot 0.7413$.

For comparison to prior work, including OpenCap, we additionally report the Mean Absolute Error following the procedure in \cite{uhlrich_opencap_2023}. This Mean Absolute Error calculation includes the following joints: pelvis rotations (3), hips (2×3), knees (2×1), ankles (2×2), and lumbar (3).

To elucidate where in the gait cycle errors occur, we extracted 1,971 gait cycles from our MMMC dataset with paired MMMC (multi-camera) and PBL (single-camera) recordings and performed a paired SPM1D t-test \cite{pataky_generalized_2010,pataky_one-dimensional_2012} between hip and knee flexion angles.   

\subsection*{Gait Deviation Index}

The GDI maps cycle-aligned joint angles to a lower-dimensional subspace in which we can measure distance from a normative population. We computed gait event timings for every self-selected gait trial using a pretrained transformer from our prior work \cite{cotton_transforming_2022}, after discarding the first and last step. This generated 13,609 gait cycles from the PBL system. Of these, 1,267 were of control participants and we use these and 4,675 gait cycles from our data collection at the American Society of Biomechanics meeting and 3,066 cycles from our MMMC system as normative data.

As PBL most accurately captures hip flexion, hip adduction, and knee flexion angles, we only used these kinematic traces in the GDI. Each kinematic trace was interpolated to 50 points at 2\% increments across the gait cycle and concatenated to a 150 $\times$ 1 column vector. We extract right-sided waveforms aligned between right initial contact events and aligned left-sided waveforms to the left foot initial contact.

While the original GDI \cite{schwartz_gait_2008} used singular value decomposition to reduce this high dimensional data, more recent findings \cite{marks_measuring_2018} demonstrate that principal component analysis (PCA) finds a more suitable subspace. We fit PCA to our 150 $\times$ 21,350 matrix, finding that 12 components accounted for 95\% of the variance. 

This resulted in a GDI value for every step a subject took. We averaged a participant's GDI over all steps (left and right) they took on a given day to compute a singular GDI number for each participant and session.

\subsection*{Statistics}
To compare population-level GDIs, we first assessed normality using the Shapiro-Wilk test \cite{shapiro_analysis_1965}. For comparisons between two normally distributed populations, we used Student’s t-test. When at least one distribution was non-Gaussian, we used the Mann-Whitney U test to assess group differences.

To evaluate relationships between smartphone-based metrics and clinical scales, we used different correlation measures based on variable type. For associations between two continuous variables (e.g., GDI and 10MWT speed), we computed the Pearson correlation coefficient. For associations between a continuous variable and an ordinal or categorical variable (e.g., GDI and mJOA), we used the Spearman rank correlation coefficient.

In addition to the Wilcoxon effect size calculated to assess responsiveness to surgical intervention, we computed the Standardized Response Mean (SRM) \cite{husted_methods_2000,liang_comparisons_1990, werner_concurrent_2020}, defined as:

\begin{equation}
SRM=\frac{\bar{D}_x}{SD(D_x)}
\end{equation}

where ${D}_x$ is the vector of pre- to post-surgery differences in measure $x$, $\bar{D}_x$ is the mean of these differences, and $SD(D_x)$ is their standard deviation. We calculated SRMs for the mJOA, cadence, double support time, and GDI.

To estimate confidence intervals for the SRM, we employed the jackknife technique \cite{miller_jackknife-review_1974} as was done in \cite{liang_comparisons_1990}. This nonparametric resampling method systematically recomputes the SRM by omitting one observation at a time, producing a set of jackknife replicates. These replicates are then used to compute pseudo-values, which approximate the influence of each observation on the overall SRM. The mean of these pseudo-values serves as a bias-corrected point estimate of the SRM, and their variability provides an estimate of its standard error. A t-distribution is then used to construct confidence intervals, enabling inference about the SRM’s precision without assuming normality of the underlying data.

We defined significance as $p<0.05$ in all cases.

\subsection*{PBL vs MMMC GDI}
To assess the impact of recording system (MMMC vs. PBL) on GDI, we conducted a separate analysis using only sessions recorded simultaneously with both systems. For each session, we computed GDI twice: once using PBL data with a PBL-based normative distribution, and once using MMMC data with an MMMC-based normative distribution. We then compared the resulting GDI values for each participant across the two systems.

\subsection*{Fitting Residuals}
We measure the 2D Fitting Residual and 3D fitting residual to evaluate the closeness of biomechanical fits to video-based joint locations. The 2D fitting residual evaluates the reconstructed model's agreement between the video-based 2D keypoints by projecting the reconstruction through a known camera model and comparing the Euclidean distance in the pixel space. The 3D Fitting Residual compares Euclidean error between 3D keypoints of the biomechanical reconstruction and video-based 3D keypoints or 3D markers.

%% file: 6supplementary_results.tex
\subsection*{Detailed Analysis of Joint Angle Errors}
Overall, Median Absolute Error of joint angles was approximately 3° across datasets, populations, and activities. Below, we highlight key trends and deviations from this general pattern, offering possible explanations and practical takeaways for users. However, we do not perform extensive statistical testing across the many variables examined, as the focus of this work is on enabling accessible motion capture rather than achieving the most precise measurement possible.

\input{tables/supplementary_joint_errors_by_dataset}

Across datasets, the BML-MoVi dataset exhibited the lowest Median Absolute Error (2.74°; Table S\ref{tab:supplementary_joint_errors_by_dataset}, Fig. \ref{fig:supplementary_histogram}C), which is expected since it included only control participants. The MMMC dataset showed similar errors when using a handheld camera (2.79°), but higher errors with a static camera (2.96°). This increase is likely due to the greater distance between the participant and camera in the static setup, whereas the handheld condition involved a researcher following the participant more closely (Fig. \ref{fig:overlay}). Some of the added error in the MMMC dataset may also be attributable to noise in the MMMC system itself, despite the use of multiple cameras.

When comparing clinical populations within the MMMC dataset, control participants showed the lowest Median Absolute Error (2.51°), while participants with neurological conditions showed the highest (3.32°; Table S\ref{tab:monocular_joint_errors_by_population}). This is consistent with expectations: many neurological participants used assistive devices—such as canes, walkers, or ankle-foot orthoses—which can obstruct joint visibility and hinder keypoint detection. Additionally, physical assistance from a therapist using a gait belt may contribute to further occlusion and error. Lower limb prosthesis users (LLPUs) also exhibited elevated Median Absolute Error (2.97°), reflecting known limitations of keypoint detectors when applied to prosthetic limbs \cite{cimorelli_validation_2024}.
\input{figures/supplementary_histogram_fig}

Across different activities, Median Absolute Error remained relatively stable (Fig. S\ref{fig:supplementary_histogram}A). Unsurprisingly, standing had the lowest Median Absolute Error (1.80°), followed by the Four Square Step Test (FSST; 2.47°), both of which involve minimal occlusion. In contrast, activities such as the Timed Up and Go (3.17°), tandem gait (3.22°), and walking (2.95°) yielded slightly higher errors—likely due to partial leg occlusions caused by the oblique handheld filming angle used during much of these trials.

Joint-level trends in Median Absolute Error were largely consistent across populations (Table S\ref{tab:monocular_joint_errors_by_population}). Notably, ankle Median Absolute Error exceeded 5° in all populations, indicating a clear need for improved foot keypoint detection. Elbow flexion error approached 5° in clinical cohorts, likely because one arm is frequently occluded during walking. Hip flexion error was higher than expected in the BML-MoVi and static-camera MMMC datasets (Table S\ref{tab:supplementary_joint_errors_by_dataset}). These datasets featured more frontal views of the participant, which—as shown in our viewpoint analysis (Fig. \ref{fig:quality}C)—can increase sagittal plane error. Interestingly, this often manifested as a consistent offset rather than an error in the joint’s range of motion. A more detailed investigation of these effects, including marker placement and viewpoint-dependent bias in keypoint detection, is left for future work.

\input{tables/supplementary_joint_errors_by_population}

\input{tables/mae_table}
\subsubsection*{Mean Absolute Error}
Mean Absolute Errors calculated with the same joints as used in OpenCap \cite{uhlrich_opencap_2023} were below 5 degrees for most activities and populations (Table S\ref{tab:opencap_mae}). TUG had the highest Mean Absolute Error at 5.17°, and the neurologic population had the highest Mean Absolute Error at 4.49°. For the most faithful comparison to OpenCap, we computed Mean Absolute Error on the control population during walking, finding 4.32 (2.06)° of error, which approaches OpenCap's reported error of 4.1° on the same population and activity.

\subsubsection*{SPM1D Comparison}
\input{figures/supplementary_spm_mmc_to_mono}
Paired SPM1D t-tests on per-step kinematics showed significant differences in knee and hip angles across nearly the entire gait cycle (Fig. S\ref{fig:supplementary_spm1d}). Given the large sample size (n = 1,971), these results are expected, and the absolute differences remain small. Hip flexion was consistently underestimated by the single-camera method, suggesting a potential pelvis-tilt bias, possibly driven by viewpoint-dependent errors (Fig. \ref{fig:quality}C), as most recordings were captured from behind. Knee flexion was also underestimated during stance and slightly overestimated near peak flexion. Despite the statistical significance, the magnitude of these errors is modest.

\input{figures/supplementary_rte}

\subsection*{Standardized Response Mean (SRM)}
\input{tables/supplementary_srm}
SRM results (Table S\ref{tab:supplemtary_srm}) aligned closely with the effect sizes reported in Table \ref{tab:gdi_cm_summary}. For the smartphone-based metrics, SRMs for pre/post-operative change were 0.34 for cadence, –0.48 for DST, and 0.45 for GDI. Patient-reported outcomes showed smaller SRMs of 0.20 (mJOA) and 0.14 (mJOA-LE). Notably, the confidence intervals for DST and GDI did not cross zero, whereas all other intervals did.

\subsection*{PBL vs MMMC GDI}
GDI values computed from PBL and MMMC recordings of the same sessions were highly consistent, despite differences in recording modality and normative datasets. We found a strong, significant, correlation between the two different modalities (0.85) demonstrating that PBL and MMMC both likely capture the same features needed for downstream gait analysis tasks (Fig. S\ref{fig:supplementary_gdi}). We note that the PBL-based GDI shows a greater change than the MMMC system, which could suggest some of the deviations in gait detected by the PBL system arise from measurement error from the PBL system; this is more impactful for greater gait impairments.

\input{figures/supplementary_gdi}

\subsection*{Fitting residuals}
Across all datasets and modalities, 2D and 3D residuals between reconstructed and detected joint locations were generally low at $<$5 pixels for 2D reprojection errors and $<$1.5 cm for 3D marker errors. Correlations between fitting residuals and Median Absolute Error were significant but low: 0.35 and 0.24 for 2D and 3D residuals, respectively (Fig. S\ref{fig:supplementary_residuals}A,C). While this suggests there may be some relationship between the closeness of our model fit to the actual joint angle accuracy, it is quite weak, at least at the levels most of our reconstructions achieved. This likely highlights the fact that the underlying computer vision algorithm does not have well calibrated confidence scores, something we aim to address in the future \cite{pierzchlewicz_platypose_2024,cotton_biomechanical_2025}. Importantly, we do not see any obvious issues with our approach generalizing to clinical settings as there were not large differences in either residual between in-lab and in-clinic cohorts (Fig. S\ref{fig:supplementary_residuals}B,D).

\input{figures/supplementary_residuals}

%% file: tables/supplementary_joint_errors_by_dataset.tex
\begin{table}
\captionsetup{name=Supplementary Table}

\centering
\caption{Median Absolute Error (median (nIQR)) for clinical and BML-MoVi datasets.}
\label{tab:supplementary_joint_errors_by_dataset}
\begin{tabular}{lcc|c}
\toprule
Dataset & \multicolumn{2}{c}{Clinical} & BML-MoVi  \\
Camera & Handheld & Static & Static \\
\midrule
Hip Flexion & 3.93 (2.35) & 5.37 (2.97) & 4.80 (2.68) \\
Hip Adduction & 1.69 (0.74) & 1.97 (0.84) & 1.68 (0.66) \\
Hip Rotation & 2.70 (1.00) & 2.73 (0.94) & 3.30 (1.22) \\
Knee Angle & 4.05 (1.70) & 3.92 (2.12) & 2.11 (0.53) \\
Ankle Angle & 5.38 (2.20) & 4.94 (1.97) & 3.55 (1.25) \\
Lumbar Extension & 3.09 (1.93) & 4.53 (3.86) & 5.30 (3.00) \\
Lumbar Bending & 1.43 (0.65) & 1.59 (0.76) & 1.64 (0.86) \\
Lumbar Rotation & 1.91 (1.11) & 1.78 (0.85) & 1.63 (0.84) \\
Neck Extension & 4.34 (4.40) & 4.06 (2.97) & 3.95 (2.94) \\
Neck Bending & 1.88 (1.23) & 2.05 (1.35) & 2.31 (1.02) \\
Neck Rotation & 3.01 (2.40) & 2.63 (1.62) & 2.85 (1.41) \\
Arm Flex & 3.63 (2.16) & 3.14 (1.74) & 2.31 (0.97) \\
Arm Add & 1.26 (0.70) & 1.94 (1.18) & 1.94 (0.67) \\
Elbow Flex & 4.22 (2.38) & 4.09 (2.06) & 3.85 (1.18) \\
All & 2.79 (0.86) & 2.96 (0.78) & 2.74 (0.50) \\
\bottomrule
\end{tabular}
\end{table}

%% file: figures/supplementary_histogram_fig.tex
\begin{figure}
\centering
\captionsetup{name=Supplementary Fig.}
\includegraphics[width=1.0\linewidth]{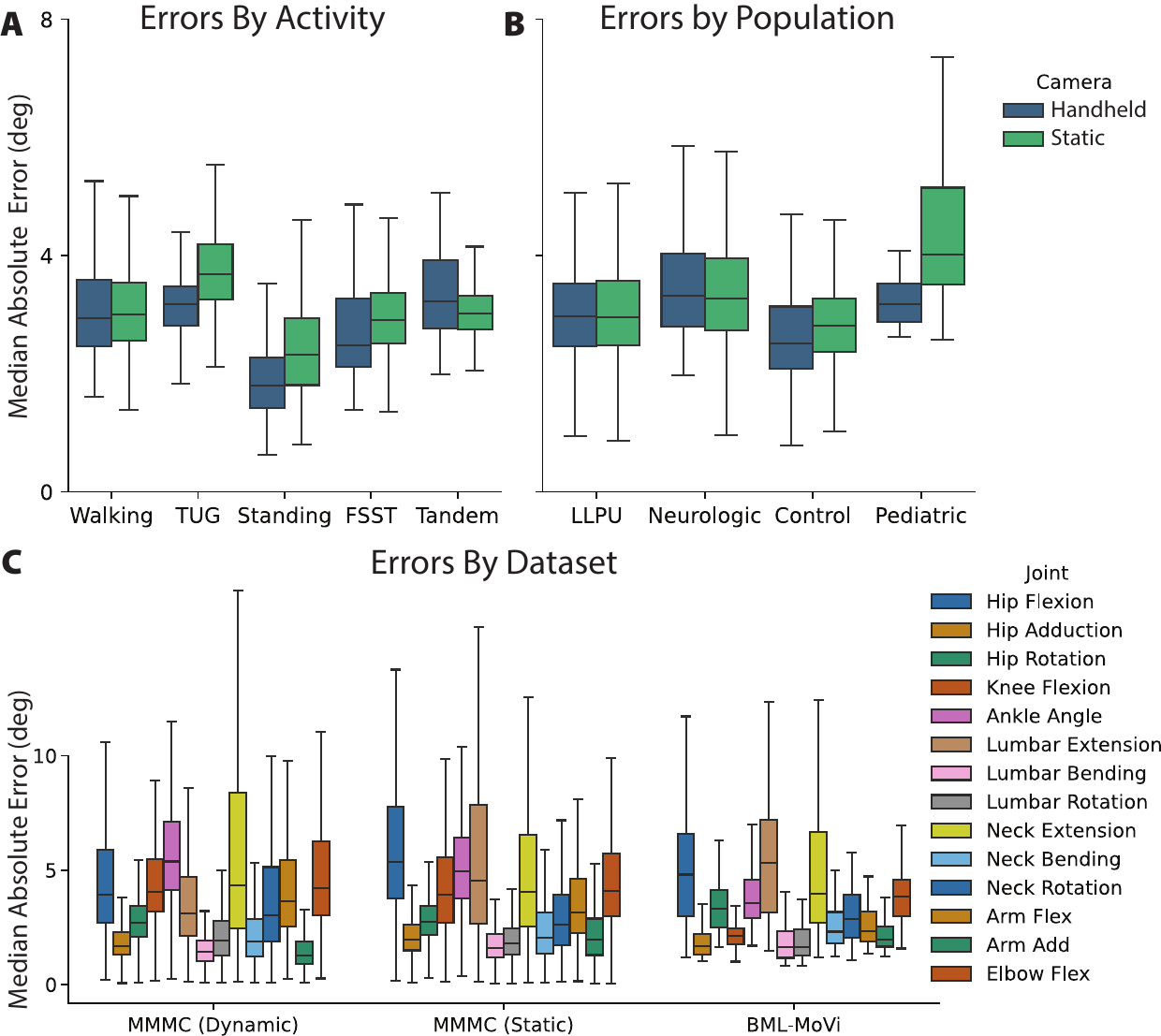}
\caption{\textbf{Joint Angle Error Distributions.} \textbf{A)} In the MMMC dataset, Median Absolute Error distributions remain relatively stable when aggregated across activity types, participant populations \textbf{(B)}, as well as between handheld (moving) and static cameras. \textbf{C)} Joint-specific Median Absolute Error distributions from both the MMMC and BML-MoVi datasets.}
\label{fig:supplementary_histogram}
\end{figure}

%% file: tables/supplementary_joint_errors_by_population.tex
\begin{table}
\captionsetup{name=Supplementary Table}
\centering
\caption{Median Absolute Error (median (nIQR)) for handheld smartphone reconstruction across different clinical populations.}
\label{tab:monocular_joint_errors_by_population}
\begin{tabular}{lcccc}
\toprule
Joint & \multicolumn{4}{c}{Median Absolute Error (nIQR)} \\
& Control & Neurological & LLPU & Pediatric \\
\midrule
Hip Flexion & 3.49 (2.27) & 3.92 (2.58) & 4.69 (2.10) & 3.06 (1.32) \\
Hip Adduction & 1.55 (0.69) & 1.93 (0.78) & 1.80 (0.80) & 1.92 (0.27) \\
Hip Rotation & 2.66 (1.17) & 2.83 (0.79) & 2.70 (0.93) & 2.74 (0.61) \\
Knee Flexion & 3.77 (1.44) & 4.36 (1.48) & 4.68 (2.08) & 3.89 (0.90) \\
Ankle Angle & 5.49 (2.75) & 5.41 (1.60) & 5.28 (1.89) & 5.30 (1.24) \\
Lumbar Extension & 3.13 (2.22) & 3.29 (1.85) & 2.89 (1.51) & 4.09 (2.28) \\
Lumbar Bending & 1.38 (0.64) & 1.58 (0.65) & 1.44 (0.66) & 1.50 (0.37) \\
Lumbar Rotation & 1.71 (1.13) & 2.19 (1.08) & 2.06 (1.02) & 1.95 (0.34) \\
Neck Extension & 3.62 (2.76) & 10.12 (6.58) & 4.27 (4.49) & 7.16 (3.60) \\
Neck Bending & 1.62 (1.12) & 2.49 (1.46) & 1.88 (1.11) & 2.37 (2.29) \\
Neck Rotation & 2.78 (2.11) & 5.00 (3.78) & 2.63 (1.90) & 3.95 (2.54) \\
Arm Flex & 3.01 (1.61) & 4.33 (3.03) & 4.43 (2.68) & 4.00 (1.47) \\
Arm Add & 1.06 (0.54) & 1.71 (1.28) & 1.47 (0.73) & 1.74 (0.95) \\
Elbow Flex & 3.59 (2.09) & 4.79 (2.47) & 4.90 (2.54) & 5.18 (2.46) \\
All & 2.51 (0.78) & 3.32 (0.92) & 2.97 (0.79) & 3.17 (0.48) \\
\bottomrule
\end{tabular}
\end{table}

%% file: tables/mae_table.tex
\begin{table}[ht]
\centering
\captionsetup{name=Supplementary Table}

\caption{Lower Body Mean Absolute Errors. Computed over trials (n = 1702) and participants (n = 107), and the reported mean is an average over activities and degrees of freedom (three for pelvis orientation, three for the lumbar, three per hip, one per knee, and two per ankle.)}
\label{tab:opencap_mae}

\begin{subtable}{\linewidth}
\centering
\caption{Across Activity}

\begin{tabular}{ccccccc}
\toprule
Activity & FSST & Standing & TUG & Tandem & Walking & All \\
\midrule
Mean Absolute Error (deg) & 4.14 (2.13) & 2.19 (1.41) & 5.17 (1.47) & 4.91 (2.67) & 4.46 (2.07) & 4.18 \\
\bottomrule
\end{tabular}

\end{subtable}

\bigskip

\begin{subtable}{\linewidth}
\centering
\caption{Across Population}
\begin{tabular}{llllll}
\toprule
Population & Control & LLPU & Neurologic & Pediatric & All \\
\midrule
Mean Absolute Error (deg) & 3.90 (1.82) & 4.29 (2.09) & 4.49 (1.94) & 4.11 (1.81) & 4.20 \\
\bottomrule
\end{tabular}
\end{subtable}
\end{table}

%% file: figures/supplementary_spm_mmc_to_mono.tex
\begin{figure}
\captionsetup{name=Supplementary Fig.}
\centering
\includegraphics[width=0.8\linewidth]{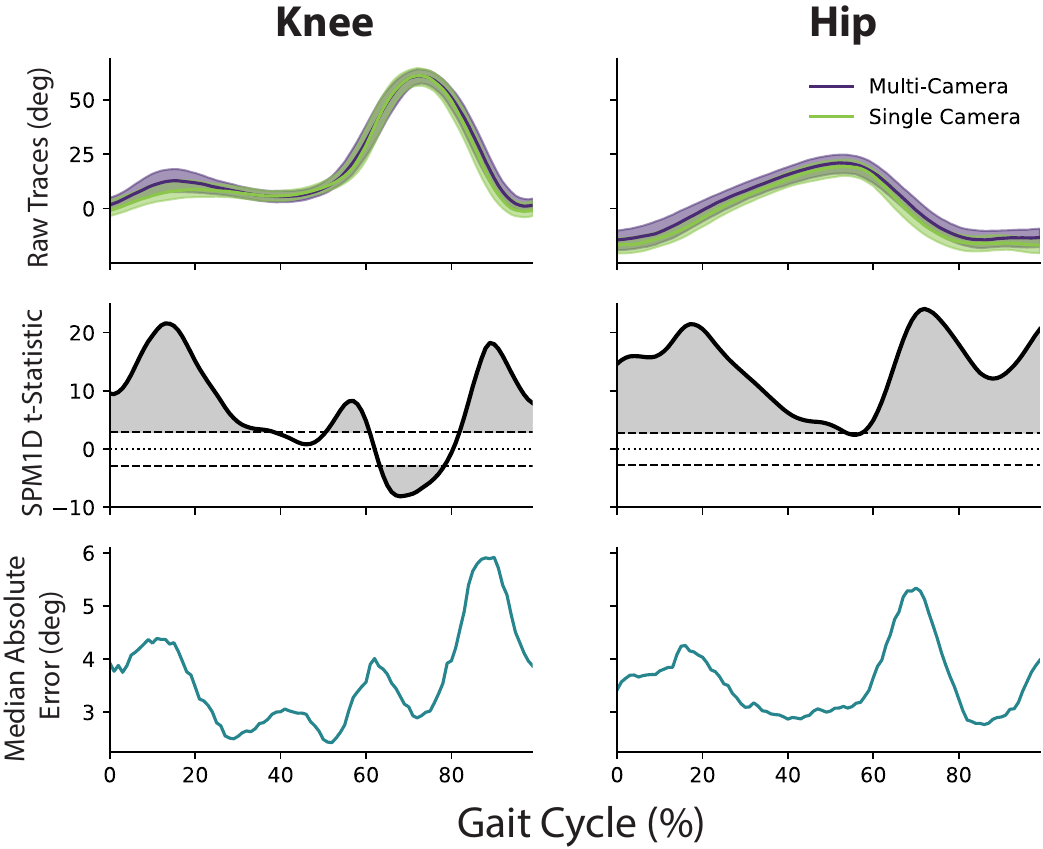}
\caption{\textbf{SPM1D Comparison of Multi Camera and Monocular.} Raw kinematic traces (n=1971) compared between the Multi-Camera (MMMMC) and Single Camera (PBL) modalities of the MMMC dataset are compared using paired SPM1D t-tests, revealing significant differences at almost every timepoint in the gait cycle. Dashed lines in second row represent p < 0.05 significance threshold.}
\label{fig:supplementary_spm1d}
\end{figure}

%% file: figures/supplementary_rte.tex
\begin{figure}
\captionsetup{name=Supplementary Fig.}
\centering
\includegraphics[width=0.8\linewidth]{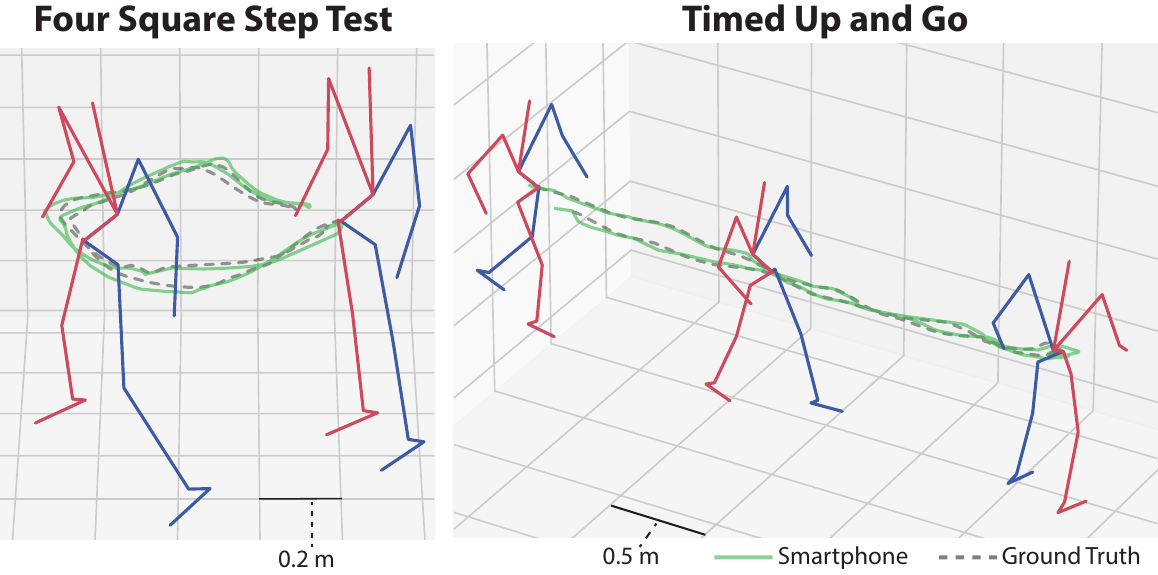}
\caption{\textbf{Root Translation Error.} Body translation in 3D space reconstructed from smartphone (green, solid) closely mirrored ground truth (grey, dashed). Note that our method accounts for smartphone rotation, but not translation.}
\label{fig:supplementary_rte}
\end{figure}

%% file: tables/supplementary_srm.tex
\begin{table}
\vspace{1em}
\centering
\captionsetup{name=Supplementary Table}
\caption{\textbf{Standardized Response Mean (SRM) due to Surgical Intervention}  SRM is a unitless measure of change in metric before and after surgery, with 95\% confidence intervals.}
\label{tab:supplemtary_srm}
\begin{tabular}{llc}
\toprule
 & Metric &  SRM \\
\midrule
\multirow{2}{*}{Patient-Reported}
 &mJOA & 0.20 (-0.38 0.78)\\
 &mJOA-LE & 0.14 (-0.45 0.72) \\
\midrule
\multirow{3}{*}{Smartphone-Based}
 &Cadence & 0.34 (-0.20 0.88) \\
 &DST & -0.48 (-0.88 -0.07) \\
 &GDI  & 0.45 (0.00 0.91)\\
\bottomrule
\end{tabular}
\end{table}

%% file: figures/supplementary_gdi.tex
\begin{figure}
\captionsetup{name=Supplementary Fig.}
\centering
\includegraphics{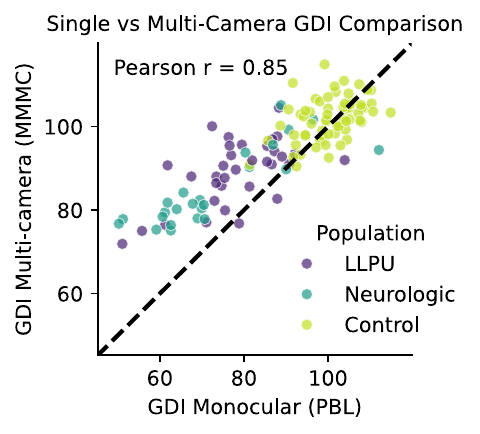}
\caption{\textbf{Gait Deviation Index Comparison between Monocular and Multi-View Fits} GDI values were computed separately using either PBL (monocular) or MMMC (multi-view) recordings from the same sessions. For each computation, we used only the data and normative distribution corresponding to the respective modality (e.g., monocular GDI was calculated using single-camera fits and a single-camera-based normative distribution).}
\label{fig:supplementary_gdi}
\end{figure}

%% file: figures/supplementary_residuals.tex
\begin{figure}
\captionsetup{name=Supplementary Fig.}
\centering
\includegraphics[width=0.8\linewidth]{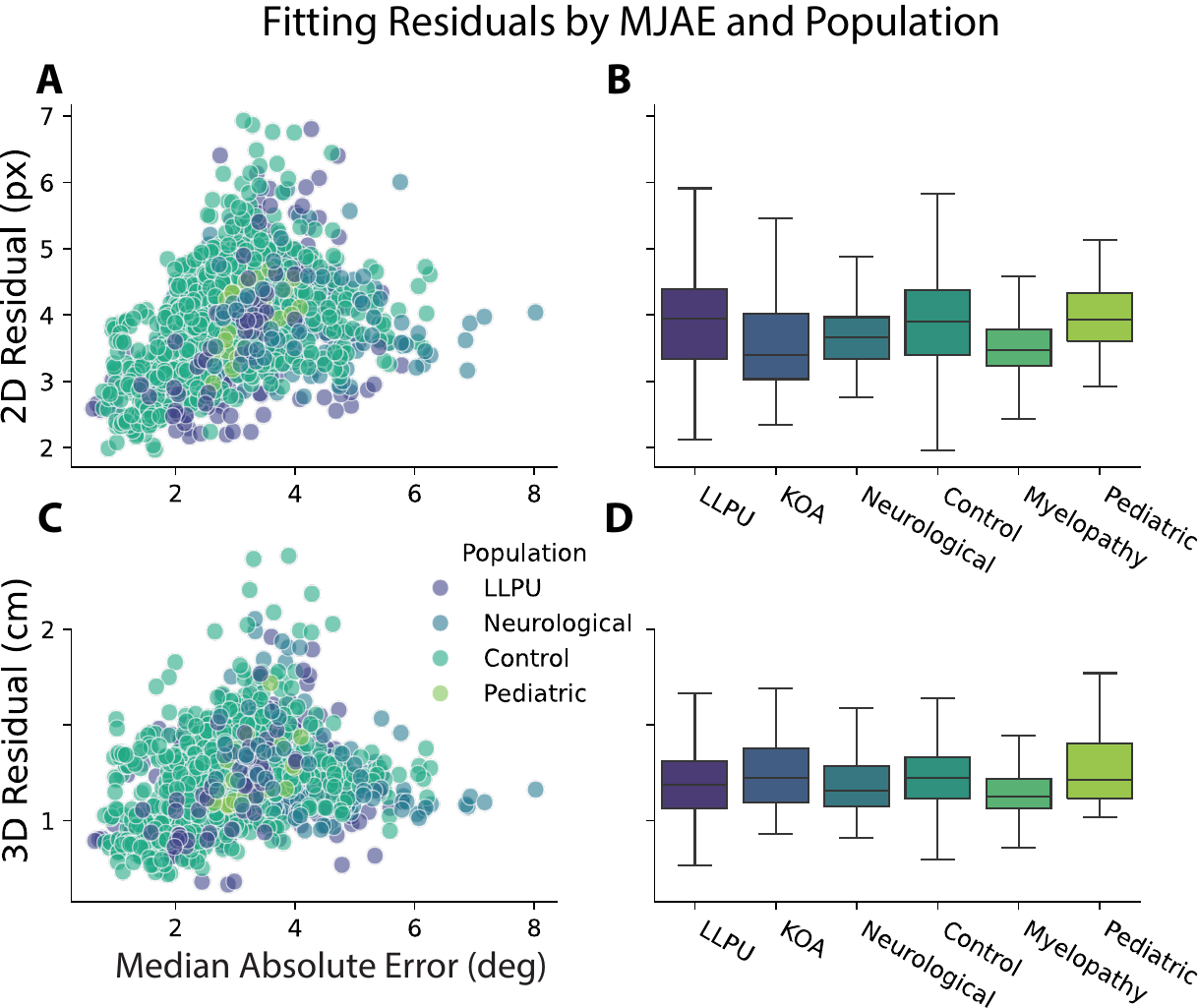}
\caption{\textbf{Fitting Residuals.} Correlation between 2D and 3D fitting residuals and Median Absolute Error is shown in (A), (C), respectively. 2D and 3D fitting residuals distributed across clinical populations is shown in (B) and (D), respectively.}
\label{fig:supplementary_residuals}
\end{figure}